\documentclass[a4paper,twoside,british]{article}

\usepackage[utf8]{inputenc}
\usepackage{babel}
\usepackage{float}
\usepackage{booktabs}
\usepackage{units}
\usepackage[hyphens]{url}
\usepackage{dcolumn}
\usepackage[]{algorithm2e}

\floatstyle{ruled}
\newfloat{algorithm}{tbp}{loa}
\providecommand{\algorithmname}{Algorithm}
\floatname{algorithm}{\protect\algorithmname}

\usepackage{epsfig}
\usepackage{subfigure}
\usepackage{calc}
\usepackage{amssymb}
\usepackage{amstext}
\usepackage{amsmath}
\usepackage{amsthm}
\usepackage{multicol}
\usepackage{tabularx}
\usepackage{pslatex}
\usepackage{apalike}
\usepackage{SCITEPRESS}     

\usepackage[unicode=true,bookmarks=true,bookmarksnumbered=false,bookmarksopen=false,breaklinks=false,backref=true,colorlinks=true,pagebackref=true]{hyperref} 

\newcommand\blfootnote[1]{%
  \begingroup
  \renewcommand\thefootnote{}\footnote{#1}%
  \addtocounter{footnote}{-1}%
  \endgroup
}

\begin{document}

\title{Rolling Shutter Camera Synchronization with Sub-millisecond Accuracy}

\author{
\authorname{Matěj Šmíd and Jiří Matas}
\affiliation{Center for Machine Perception, Czech Technical University in Prague, Czech Republic} 
\email{\{smidm, matas\}@cmp.felk.cvut.cz}
}


\abstract{A simple method for synchronization of video streams with a precision better than one millisecond is proposed. The method is applicable to any number of rolling shutter cameras and when a few photographic flashes or other abrupt lighting changes are present in the video. The approach exploits the rolling shutter sensor property that every sensor row starts its exposure with a small delay after the onset of the previous row. The cameras may have different frame rates and resolutions, and need not have overlapping fields of view. The method was validated on five minutes of four streams from an ice hockey match. The found transformation maps events visible in all cameras to a reference time with a standard deviation of the temporal error in the range of 0.3 to 0.5 milliseconds. The quality of the synchronization is demonstrated on temporally and spatially overlapping images of a fast moving puck observed in two cameras.}

\onecolumn \maketitle \normalsize \vfill

\section{INTRODUCTION}

\blfootnote{The research reported in this paper has been partly supported by the Austrian Ministry for Transport, Innovation and Technology, the Federal Ministry of Science, Research and Economy, and the Province of Upper Austria in the frame of the COMET center SCCH.}
\noindent Multi-camera systems are widely used in motion capture, stereo vision,
3D reconstruction, surveillance and sports tracking. With smartphones
ubiquitous now, events are frequently captured by multiple devices.
Many multi-view algorithms assume temporal synchronization. The problem
of multiple video synchronization is often solved by triggering the
cameras by a shared signal. This solution has disadvantages: it is
costly and might put a restriction on the distance of the cameras.
Cheaper cameras and smartphones do not have a hardware trigger input
at all. 

\begin{figure}[h]
\centering{}\includegraphics[width=1\columnwidth]{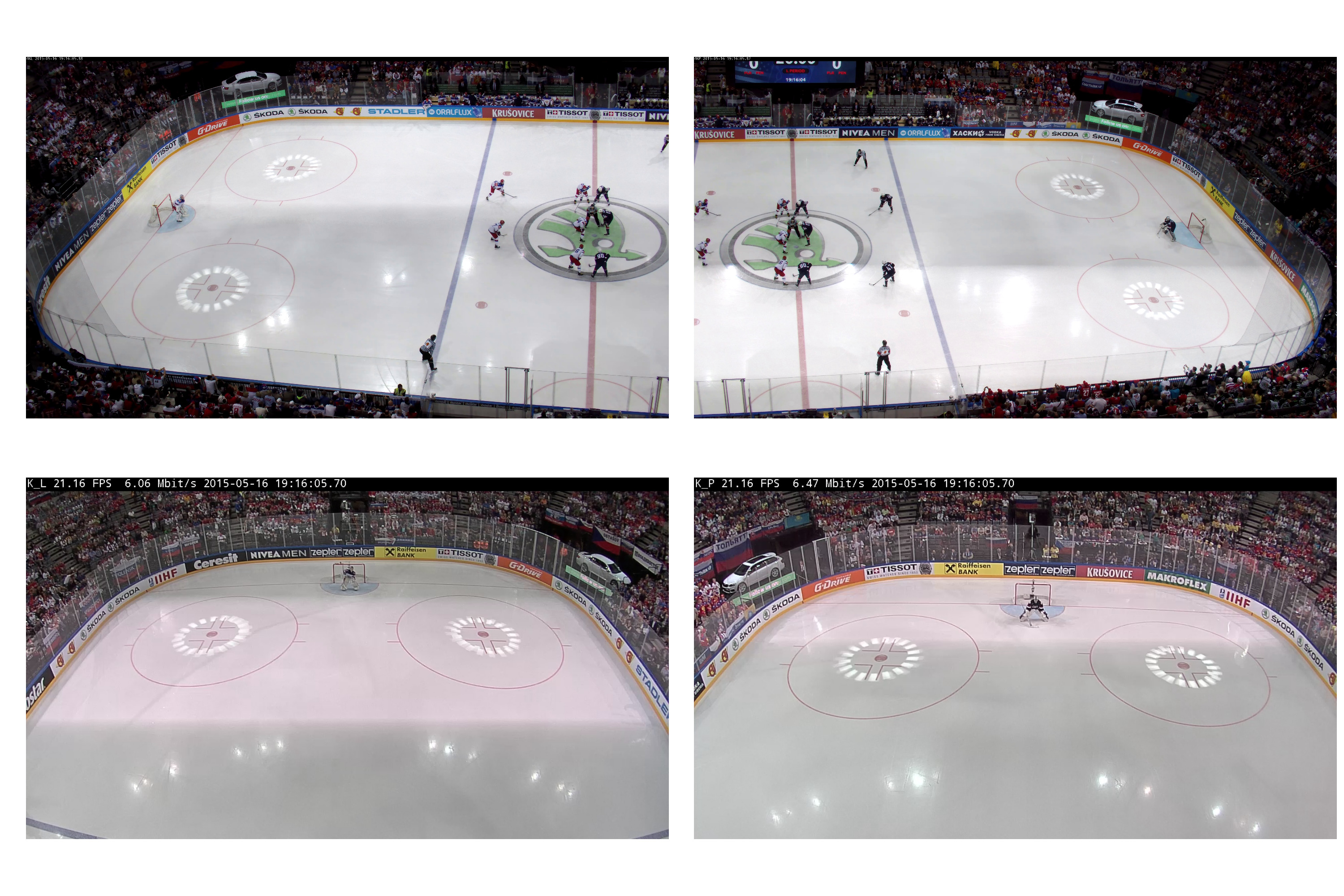}\caption{\label{fig:4flash_artefacts}Four cameras with rolling
shutter sensors capturing a scene when a photographic flash was fired. Part of  image rows integrated light from the flash. The leading and trailing edges are easily detectable and on the ice rink also clearly visible. The edges serve as very precise synchronization points. }
\end{figure}

Content-based synchronization can be performed offline and places no requirements on the data acquisition. It has received stable attention in the last
20 years \cite{Stein1999,Caspi2002,Tresadern2003,Lei2006,Padua2010}.
Some of the methods require calibrated
cameras, trackable objects, laboratory setting or are limited to
two cameras. The wast majority of the methods requires overlapped views. For analysis of high-speed phenomena, a very precise
synchronization is critical. The problem of precise sub-frame synchronization
was addressed in \cite{Caspi2006,Tresadern2009,Dai2006}.

We propose a very simple yet sub-millisecond accurate method for video
data with abrupt lighting changes captured by rolling shutter cameras.
Such lighting changes could be induced for example by photographic
flashes, creative lighting on cultural events or simply by turning on a light source. In controlled conditions,
it is easy to produce necessary lighting changes with a stock camera
flash. 

It is very likely that an existing multi-view imaging system uses
rolling shutter sensors or that a set of multi-view videos from the
public was captured by rolling shutter cameras. The expected image
sensor shipment share for CMOS in 2015 was 97\% \cite{IHSInc2012}.
Most of the CMOS sensors are equipped with the rolling shutter image
capture.

The proposed method assumptions are limited to:
\begin{itemize}
\item a few abrupt lighting changes affecting most of the observed scene,
and
\item cameras with rolling shutter sensors.
\end{itemize}

The method does not require an overlapping field of view and the cameras
can be heterogeneous with different frame rates and resolutions. The
proposed method works with frame timestamps instead of frame 
numbers. This means that the method is robust to dropped frames.

When a lighting abruptly changes during a rolling shutter frame exposure, the transition edge can be reliably detected in multiple cameras
and used as a sub-frame synchronization point. An example of captured frames with an abrupt lighting change caused by a single photographic flash is shown in Figure~\ref{fig:4flash_artefacts}. 

Let us illustrate the importance of precise sub-frame synchronization
on an example of tracking ice hockey players and a puck. Players can
quickly reach a speed of $\unitfrac[7]{m}{s}$ \cite{Farlinger2007}
and the puck $\unitfrac[36]{m}{s}$ \cite{Worobets2006}. When we
consider $\unit[25]{fps}$ frame rate, a player can travel $\unit[28]{cm}$,
and a puck can move $\unit[1.44]{m}$ in the $\unit[40]{ms}$ duration
of one frame. When a synchronization is accurate up to whole frames, the mentioned uncertainties can lead to poor multi-view tracking performance.

\begin{figure*}
\centering{}\includegraphics[width=1\textwidth]{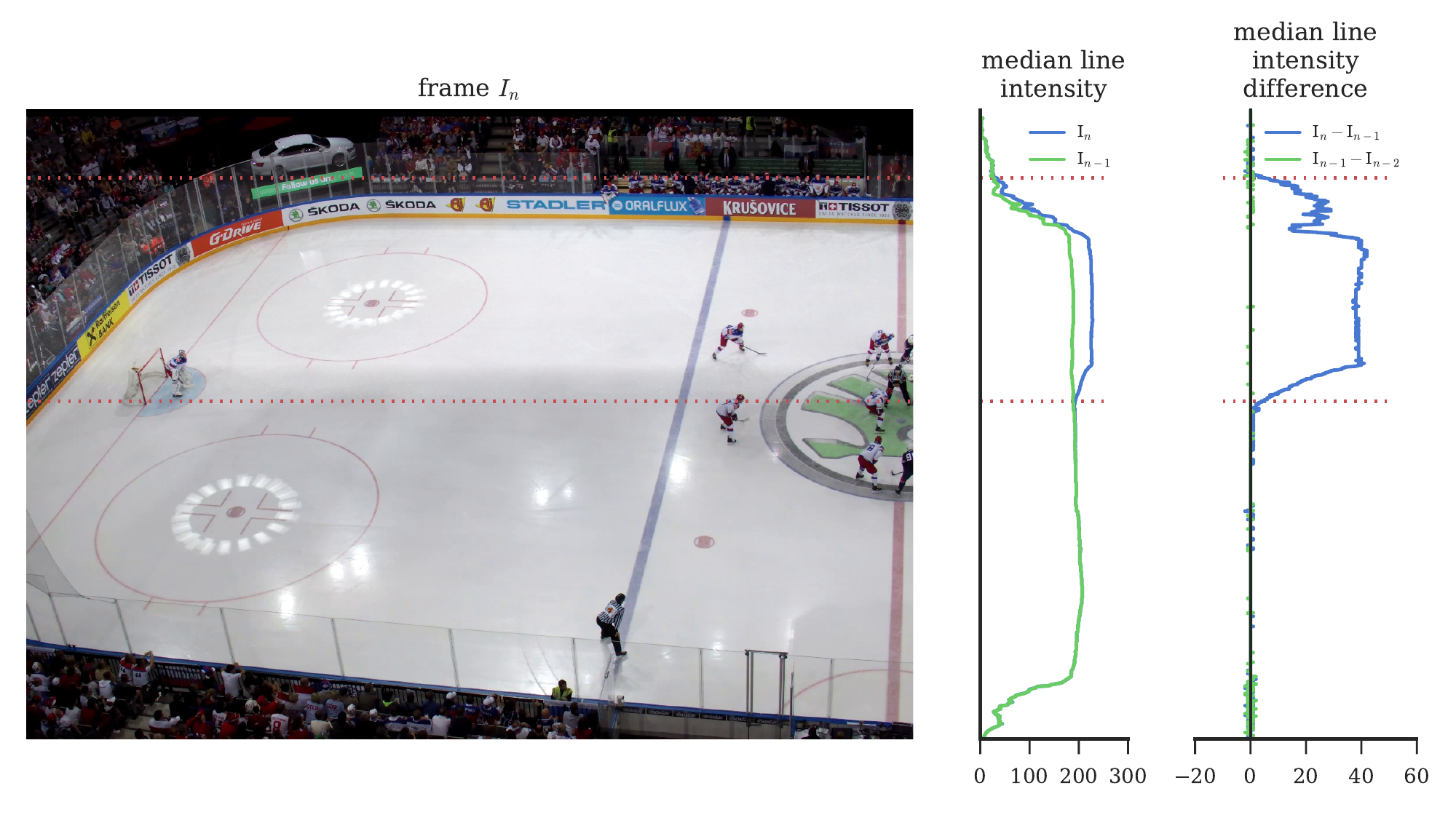}\caption{\label{fig:frame_with_profiles}Detection of an abrupt lighting change.
A photographic flash fired during the acquisition of the frame $I_{n}$. The
flash duration is shorter than the frame duration. Only the
lines that were integrating light when the flash was illuminating the scene were affected.
The red dotted lines mark the leading and trailing edges of the bright region.
The profiles on the right show pixel intensity changes in the frame
before the abrupt change and in the frame with the change. }
\end{figure*}

\section{RELATED WORK}

\noindent The computer vision community keeps stable attention to the video
synchronization problem. The issue was approached in multiple directions.
Synchronization at the acquisition time is done either by a special hardware
or using a computer network to synchronize time or directly trigger cameras. A more
general approach is video content-based synchronization. The advantage
is that it does not have special acquisition requirements. We already
mentioned a number of content-based methods. We will review works that make use
of a rolling shutter sensor or photographic flashes which are the most relevant to
our method.

\cite{Wilburn2004} construct rolling shutter camera array that was
able to acquire images at 1560 fps. The cameras were hardware synchronized,
and the rolling shutter effect was mitigated by outputting slices of
a spatio-temporal image volume. 

\cite{Bradley2009} approach the rolling shutter image capture in
two ways. First, they acquire images with stroboscopic light in a
laboratory setting, and extract and merge only rows affected by a
light pulse that possibly span over two consecutive frames. By changing
the frequency and duration of the flashes they effectively create
a virtual exposure time and a virtual frame rate. Second investigated
approach merges two consecutive frames by a weighted warping along optical
flow vectors. This is similar to the spatio-temporal method.

Cinematography focused methods for the rolling shutter sensor acquisition
were studied by \cite{Hudon}. They analyse stroboscopic light artefacts
for the purpose of image reconstruction. 

\cite{Atcheson2008} applied the rolling shutter flash based synchronization
and spatio-temporal volume slice approach to capture gas flows for
a 3D reconstruction.

The use of photographic flashes for synchronization appeared to our
knowledge first in \cite{Shrestha2006}. They find a translation between
two video sequences by matching sequences of detected flashes. The
final synchronization is accurate to the whole frames.

None of the rolling shutter or flash based approaches known to us pays attention
to dropped frames and a camera clock drift.

\section{METHOD}

\begin{figure}
\centering{}\includegraphics[width=1\columnwidth]{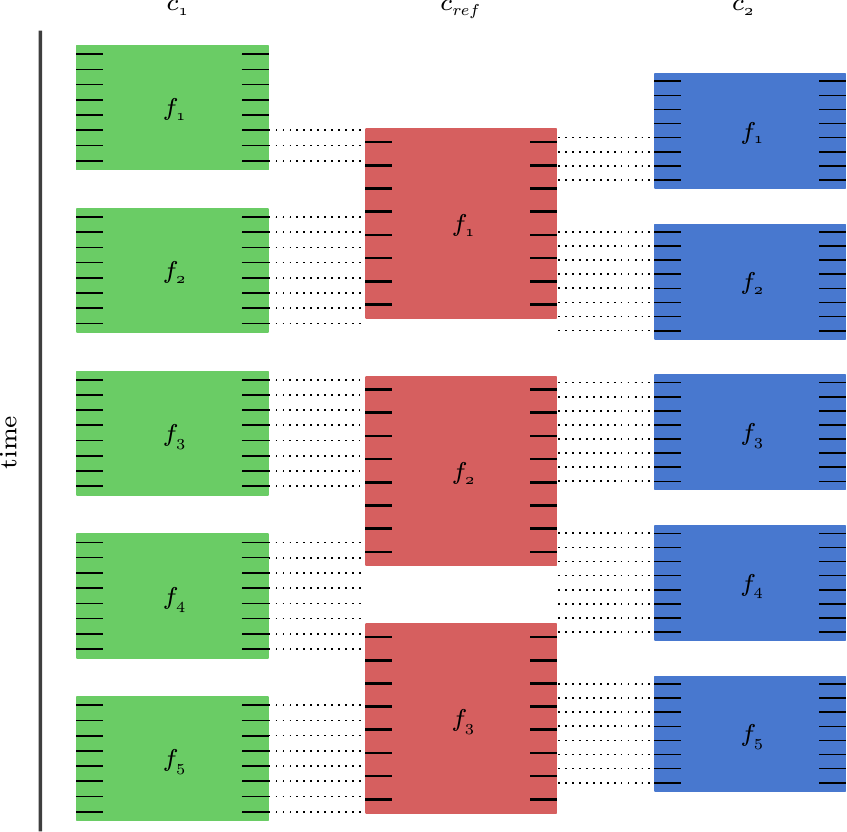}\caption{\label{fig:synchronization}Sub-frame synchronization of the cameras
$c_{1}$ and $c_{2}$ with respect to the reference camera $c_{\mathrm{ref}}$.
Frame rates, resolution and temporal shifts between cameras differ.
The short black lines on the sides of frame rectangles represent image
rows. We find an affine transformation $s^{c}(f,r)\rightarrow t^{\mathrm{ref}}$
for every camera $c$ that maps a time point specified by a frame number $f$
and a row number $r$ to the reference camera time $t^{\mathrm{ref}}$.
The dotted lines show a mapping of time instants when rows in $c_1$ and $c_2$ are captured to the reference camera time. }
\end{figure}

\noindent The inputs for the synchronization algorithm are frame timestamps
extracted from video files or network streams and detected transition edges of abrupt lighting changes. We will refer to the transition edges as \textit{synchronization events} or simply \textit{events}. We find synchronization transformations $s^{c}(f,r)\rightarrow t^{\mathrm{ref}}$ for all cameras $c$ (except a reference camera $c_{\mathrm{ref}}$)
that map each camera temporal position $(f, r)$ to the reference camera time
$t^{\mathrm{ref}}$. The temporal position is defined by a frame, row pair $\left(f,r\right)$. The situation is presented in Figure~\ref{fig:synchronization}.

To correctly model the sub-frame accurate synchronization transformation
we have to take into account missing frames, different frame rates,
a drift of image sensors clock and hidden \emph{dark
rows} in image sensors.

\subsection{On Time Codes}

An ideal timing of video frames assumes a stable frame rate $\mathrm{fps}$
and no skipped frames. Time of the first row exposure of a frame $i$ is then $t(i)=i\cdot\frac{1}{\mathrm{fps}}+t_{0}, i \in \{0, 1, \dots\}$.
Unfortunately, this is not true for most of the real-world video sequences.
The most common deviation from the ideal timing is a dropped frame caused by high CPU load on the encoding system. When a frame is not encoded before the next one is
ready, it has to be discarded. Almost all video sources provide frame timestamps or frame durations. This information is necessary to maintain very precise synchronization over tenths of minutes. We'll briefly present frame timing extraction from container format MP4 and  streaming protocol RTP.

Video container files encapsulate image data compressed by a video codec.
The timing data is stored in the container metadata. The MP4\footnote{officially named MPEG-4 Part 14} file format is based on Apple
QuickTime. Frame time-stamps are encoded in \emph{Duration}
and \emph{Time Scale Unit} entries. The \emph{Time Scale Unit} is defined as ``the number of time units
that pass per second in its time coordinate system''. 

A frequent streaming protocol is Real Time Transfer Protocol (RTP).
The codec compressed video data is split into chunks and sent typically
over UDP to a receiver. Every packet has an RTP header where the
\emph{Timestamp} entry defines the time of the first frame in the
packet in units specific to a carried payload: video, audio or other.
For video payloads, the \emph{Timestamp} frequency is set to $\unit[90]{kHz}$.

\subsection{Rolling Shutter\label{subsec:Rolling-Shutter}}

Historically, cameras were equipped with various shutter systems.
To name the mechanical shutters, prevalent were the focal plane shutters -
where two curtains move in one direction or the diaphragm shutters where
a number of thin blades uncover circular aperture. The electronic shutters
implemented in image sensors are either global or rolling. CCD type
image sensors are equipped with a global shutter, but are already
being phased out of the market. Most of the CMOS sensors have a rolling
shutter. Recently, a global shutter for the CMOS sensors was introduced,
but consumer products are still rare. 

All shutter types except the global shutter exhibit some sort of image
distortion. Mostly different regions of the sensor (or film) integrate light 
in a different time or the exposure time differs. 

\begin{figure}
\centering{}\includegraphics[width=1\columnwidth]{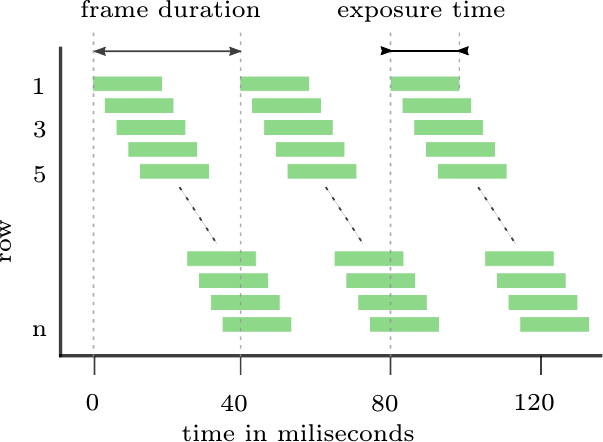}\caption{\label{fig:rolling_shutter}The figure illustrates rows exposure in time. The green rectangles represent the time spans when a row is integrating light. In rolling shutter sensors the rows do not start to integrate light at the same time. Instead the integration begins sequentially with small delays. }
\end{figure}

The rolling shutter equipped image sensor integrates light into the pixel
rows sequentially. In the CMOS sensor with the rolling shutter, an electrical
charge integrated in all pixels can not be read at once.
The readout has to be done row by row. For illustration see Figure~\ref{fig:rolling_shutter}.
To preserve constant exposure time for all pixels on the sensor, the exposure starts has to be sequential exactly as the readouts are. This means that every row captures
the imaged scene in a slightly different moment. Typically a majority
of the row exposure time is shared by spatially close rows \cite{ONSemiconductor:MT9P031,Sony:IMX322LQJ-C}.

\begin{figure}
\centering{}\includegraphics[width=1\columnwidth]{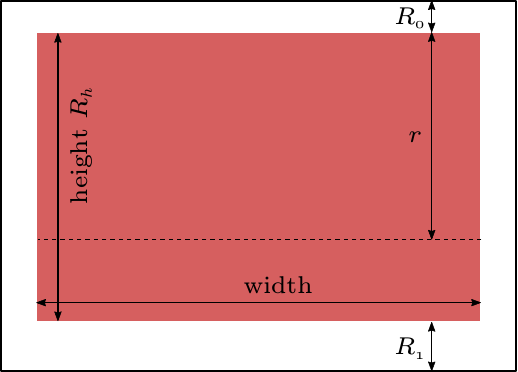}\caption{\label{fig:hidden_pixels}The active pixel matrix on image sensor
is surrounded by a strip of hidden pixels, sometimes also called ``dark'' pixels. They serve
for black colour calibration or to avoid edge effects when processing
colour information stored in the Bayer pattern \cite{ONSemiconductor:MT9P031}.
The rolling shutter model (Equation \ref{eq:subframe_timing}) assigns a sub-frame time to a row $r$.}
\end{figure}

To properly compute the start time of a row exposure we have to take into account 
hidden pixels around the active pixel area. The most image sensors use the hidden
pixels to reduce noise and fix colour interpretation at the sensor edges (Figure~\ref{fig:hidden_pixels}).
This means that there is a delay, proportional to $R_{0}+R_{1}$, between reading out the last row of a frame and the first row of the next one. Camera
or image sensor specifications often include \textit{total} and \textit{effective} pixel
count. The difference between the two values is the number of hidden
pixels.

Now it is straightforward to compute sub-frame time for a frame $f$ and
a row $r$ as
\begin{equation}
t(f,r)=t_{f}+\frac{R_{0}+r}{R_{0}+R_{h}+R_{1}}\cdot T_{\mathrm{frame}},\label{eq:subframe_timing}
\end{equation}
where $R_{0},$$R_{h}$, $R_{1}$ are row counts specified in Figure~\ref{fig:hidden_pixels},
$t_{f}$ is the frame timestamp and $T_{\mathrm{frame}}$ is the nominal
frame duration. The constants $R_{0}$ and $R_{1}$ can
be found in the image sensor datasheet or the summary value $R_{0}+R_{h}+R_{1}$
of total sensor lines can be estimated, as demonstrated in Subsection \ref{subsec:Synchronization}. 

\subsection{Abrupt Lighting Changes\label{subsec:Abrupt-Light-Changes}}

Abrupt lighting changes are trivially detectable and are suitable for
sub-frame synchronization with rolling shutter sensors.

The only requirement is that the majority of the observed scene 
receives light from the source. Many multi-view recordings already
fulfil the requirement. Professional sports photographers commonly
use flashes mounted on sports arena catwalks to capture photos during
indoor matches\footnote{\url{http://www2.ljworld.com/news/2010/mar/21/behind-lens-story-behind-those-flashing-lights-nca/}},
mobile phones or DSLRs flashes are used at many social occasions
that are recorded. Creative rapidly changing lighting is frequent at cultural
events such as concerts.

\begin{figure}
\centering{}\includegraphics[width=1\columnwidth]{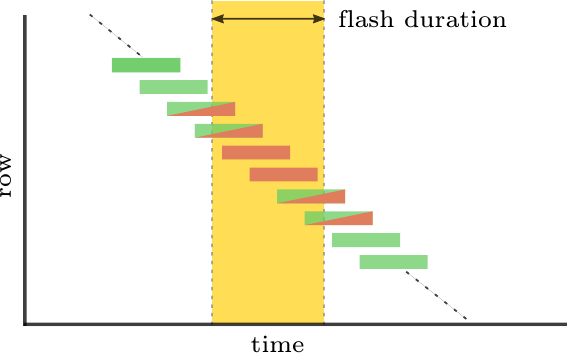}\caption{\label{fig:rolling_shutter_event}Short abrupt lighting event, e.g.,
photographic flash, affects only part of the frame rows, in red colour,
due to the rolling shutter capture. Rows, depicted half
filled in green and red, are being captured at the time of the lighting change. Such rows integrate light of the lighting event only partially. The longer the exposure time the more rows capture an onset of an event.}
\end{figure}

For photographic flashes (Figures ~\ref{fig:4flash_artefacts}, \ref{fig:rolling_shutter_event}), it is possible to detect both leading and trailing
edges. A flash duration is typically one order of magnitude
shorter than a frame duration. Flashes produce light for $\nicefrac{1}{1000}$
to $\nicefrac{1}{200}$ of a second in contrast to $\unit[40]{ms}$
frame duration of a $\unit[25]{fps}$ recording. 

\begin{figure}
\centering{}\includegraphics[width=1\columnwidth]{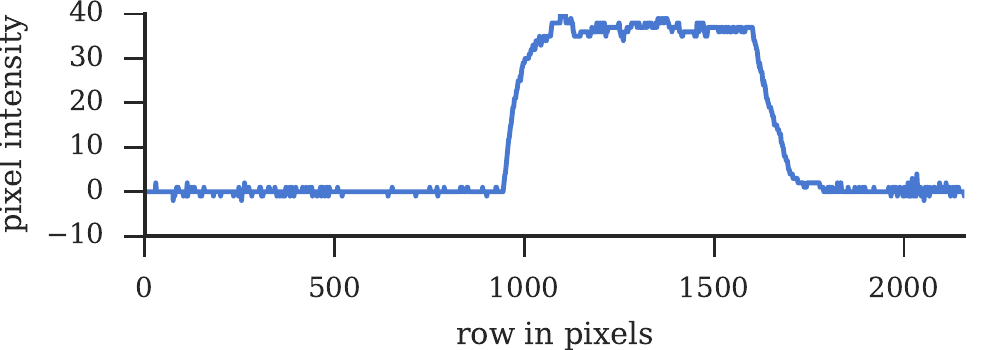}\caption{\label{fig:diff_profile}Median line intensity difference between consecutive frames in a moment of a flash. Rows in range 950-1700 were captured when the photographic flash has illuminated the scene. An
exponential character of the leading and trailing edges is related
to the physical process of the capacitor discharge in a flashtube.}
\end{figure}

An example profile of the captured light intensity by a rolling shutter
sensor is in Figure \ref{fig:diff_profile}. The shape of the profile
is formed by two processes. The exponential form of the transition edges corresponds
to the physical properties of the lighting source. The partially affected
rows at the start and end of an event contribute with a linear ramp
to the profile shape. 

\begin{figure}
\centering{}\includegraphics[width=1\columnwidth]{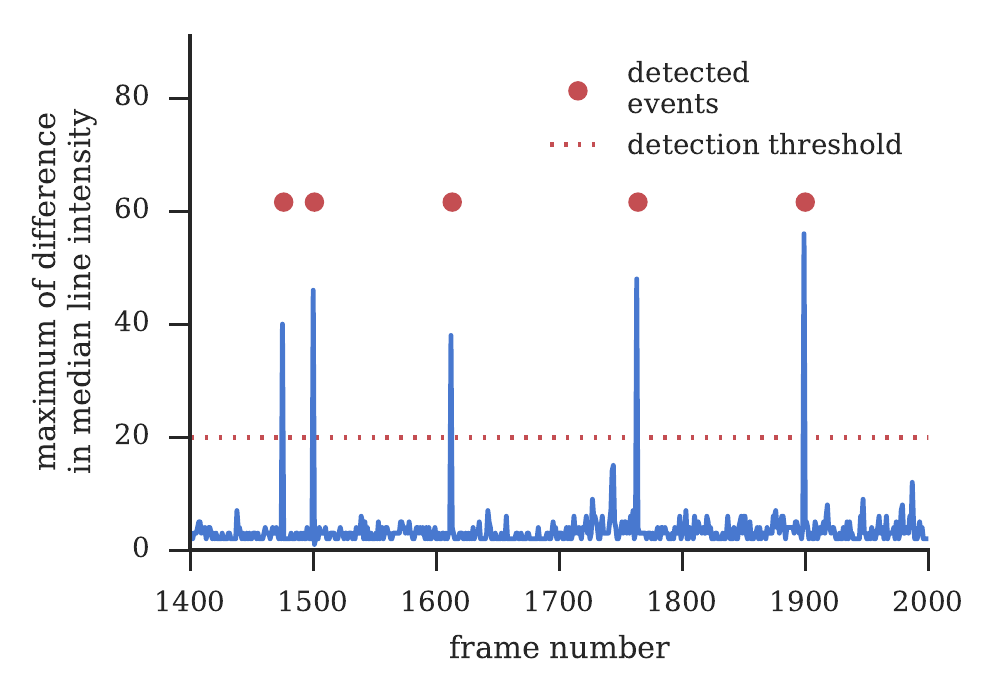}\caption{\label{fig:event_detection}Detection of the abrupt lighting changes. A median line intensity profile is computed for every frame. Then the profiles in consecutive frames are subtracted. The difference maxima for range of frames is plotted above. The clearly visible peaks correspond to the lighting changes. We threshold the values and detect the events marked on the plot by the red dots.}
\end{figure}

The detection of the abrupt lighting changes is robust and straightforward.
As we require that the lighting affects most of the scene, the maximum
of difference of median line intensity for a frame shows distinct peaks, see Figure~
\ref{fig:event_detection}. We simply threshold the values to get
the frames with the \textit{events}. We use the leading edge as the \textit{synchronization event}. The \textit{event} row is found in the differences of median line intensity profiles, see Figure~\ref{fig:frame_with_profiles}. The method is summarized in Algorithm \ref{alg:event_detection}. 

\begin{algorithm}
	\SetKw{KwInSet}{in}
	\SetKw{KwWhere}{where}
	\KwIn{image sequences}
	\KwOut{synchronization events}
	\ForEach{camera}{
		\ForEach{frame}{
			$m_f$ := line median intensity (frame) \;
			
			$m_f \in \mathbb{N}^n$, where $n$ is frame height
		}
		\ForEach(compute difference profiles){frame}{
			$d_f:=m_{f}-m_{f-1}$ \;
			
			$d_f \in \mathbb{Z}^n$, where $n$ is frame height
		}				
		\For{f \KwInSet $ \{ f \mid  \max(d_f)>\mathrm{threshold} \}$}{			
			$r$ := find raising edge row in $d_f$ \;
			
			event := $(f, r)$ \;
		}
	}
%
	\KwRet{events}
	\caption{\label{alg:event_detection}Detection of synchronization events}
\end{algorithm}

\subsection{Synchronization\label{subsec:Synchronization}}

We model the time transformation $s^{c}(f,r)\rightarrow t^{\mathrm{ref}}$ from a camera $c$ to a reference camera $c_{\mathrm{ref}}$ as~an~affine mapping similar to \cite{Padua2010}. Substantial difference is that we operate on timestamps instead of frame numbers. The transformation maps the \textit{events} detected in camera
$c$ to the same \textit{events} in the time of a~reference camera $c_{\mathrm{ref}}$.
The dominant part of the transformation $s^{c}(f,r)$ is a temporal
shift between cameras $c$ and $c_{\mathrm{ref}}$. The synchronization
model consisting of a constant temporal shift is usable only for shorter
sequences. We found out in experiments that camera clocks maintain
stable frame duration, but the reported time units are not precisely equal. This deficiency is known as the clock drift. We compensate the drift by a linear component of the transformation. 

The proposed transformation is
\begin{equation}\label{eq:transformation}
s(f,r;\alpha,\beta)=\alpha t_{f}+\beta+r\cdot\frac{T_{\mathrm{frame}}}{R},
\end{equation}
where $\alpha$ is the camera clock drift compensation, $\beta$ is the temporal
shift, $f$ is the frame number, $r$ is the row number, $t_{f}$ is the frame
acquisition timestamp and $R = R_{0}+R_{h}+R_{1}$ is the total number of sensor rows.

The goal of the synchronization is to find $s^{c}(f,r;\alpha^{c},\beta^{c})$
for all cameras in $C=\left\{ c_{1},c_{2},\ldots,c_{n}\right\} $
except for a reference camera $c_{\mathrm{ref}}$. 

For an event observed in camera $c$ and $c_{\mathrm{ref}}$ at $\left(f^{c},r^{c}\right)$
and $\left(f^{c_{\mathrm{ref}}},r^{c_{\mathrm{ref}}}\right)$ the synchronized
camera time and the reference camera time should be equal: 
\begin{equation}
s^{c}(f^{c},r^{c};\alpha^{c},\beta^{c})=t^{c_{\mathrm{ref}}}(f^{c_{\mathrm{ref}}},r^{c_{\mathrm{ref}}}).\label{eq:synch_eq_reference}
\end{equation}

We have demonstrated how to detect abrupt lighting changes in Subsection
\ref{subsec:Abrupt-Light-Changes}. In the next step, we manually align time
in cameras $c$ and $c_{\mathrm{ref}}$ up to whole frames, e.g., for the first matching
event, and automatically match the rest of the events to get:

\begin{eqnarray}
E^{c,c_{\mathrm{ref}}} & = & \bigg\{\left\{ \left(f_{1}^{c},r_{1}^{c}\right),\left(f_{1}^{c_{\mathrm{ref}}},r_{1}^{c_{\mathrm{ref}}}\right)\right\} \label{eq:matching_events}\\
& ,..., & \left\{ \left(f_{k}^{c},r_{k}^{c}\right),\left(f_{k}^{c_{\mathrm{ref}}},r_{k}^{c_{\mathrm{ref}}}\right)\right\} \bigg\}.\nonumber 
\end{eqnarray}

Now we can construct overdetermined
system of Equations \ref{eq:synch_eq_reference} for $k$ pairs of
matching events $E^{c,c_{\mathrm{ref}}}$. The least squares solution gives the unknowns $\alpha^{c},\beta^{c}$. Optionally also the sensors properties $T^{c}_{\mathrm{row}}:=\nicefrac{T^{c}_{\mathrm{frame}}}{R^{c}}$ and $T^{c_{\mathrm{ref}}}_{\mathrm{row}}:=\nicefrac{T^{c_{\mathrm{ref}}}_{\mathrm{frame}}}{R^{c_{\mathrm{ref}}}}$ can be estimated, when these are not available in the image sensors datasheets.

When synchronizing more than two cameras, one system of equations for all cameras has to be constructed to estimate the reference camera \textit{time per image row} $T^{c_{\mathrm{ref}}}_{\mathrm{row}}$ jointly.

We summarize the synchronization process in Algorithm \ref{alg:synchronization}. The single global time for a frame $f$ and row $r$ is computed using Equation \ref{eq:subframe_timing} for a reference camera and using Equation \ref{eq:transformation} for other cameras.

\begin{algorithm}
	\SetKw{KwInSet}{in}
	\SetKw{KwWhere}{where}
	\KwIn{frame timestamps, detected synchronization events, reference camera $c_{\mathrm{ref}}$}
	\KwOut{synchronization parameters}
	\ForEach{$\{ c \in C \mid c \neq c_{\mathrm{ref}} \}$}{
		$E^{c,c_{\mathrm{ref}}}$ := match events in $c$ and $c_{\mathrm{ref}}$ \;
		
		\ForEach{event \KwInSet $E^{c,c_{\mathrm{ref}}}$}{
			$\left\{ \left(f^{c},r^{c}\right),\left(f^{c_{\mathrm{ref}}},r^{c_{\mathrm{ref}}}\right)\right\} := \mathrm{event} $ \;
			
			$t^c := $ time stamp for frame $f^{c}$ \;
			
			$t^{c_{\mathrm{ref}}} := $ time stamp for frame $f^{c_{\mathrm{ref}}}$
							
			add equation:
			
			$\alpha^c t^c+\beta^c + r^c \cdot T^c_{\mathrm{row}}
			= t^{c_{\mathrm{ref}}}+r^{c_{\mathrm{ref}}}\cdot T^{c_{\mathrm{ref}}}_{\mathrm{row}}$ \;
			
			to the system of equations \;								
		}
	}
	solve the system in a least squares sense \;
	
	\KwRet{$\{\alpha^{c},\beta^{c}, T^{c}_{\mathrm{row}} \mid c \in C $ $\mathrm{and} $ $c \neq c_{\mathrm{ref}} \}, T^{c_{\mathrm{ref}}}_{\mathrm{row}}$}
	\caption{\label{alg:synchronization}Multi-camera Synchronization}
\end{algorithm}

\section{DATA\label{sec:DATA}}

\noindent The ice hockey data consists of one complete USA versus Russia match
captured by 4 cameras. The company Amden s.r.o. provided us the data recorded
on the International Ice Hockey Federation World Championship 2015.
Example images from the cameras are on Figure~\ref{fig:4flash_artefacts}.
The cameras 1 and 2 are observing the ice rink from sides, the cameras 3 and
4 are focusing on the defending and attacking zones, that is from
the blue lines to the ends of the rink. The camera pairs 1 and 2,
and 3 and 4 are identical models with the same lenses. The cameras 1 and 2
use camera model Axis P1428E with resolution $\unit[3840\times2160]{px}$,
the cameras 3 and 4 are equipped with camera model Axis P1354 with resolution $\unit[1280\times720]{px}$.

The data was delivered in the Matroska file format and later converted
to mp4. The frame timestamps were extracted using \texttt{ffprobe}
command line utility included in the ffmpeg package.

\section{EXPERIMENTS}

\begin{figure}
\centering{}\includegraphics[width=1\columnwidth]{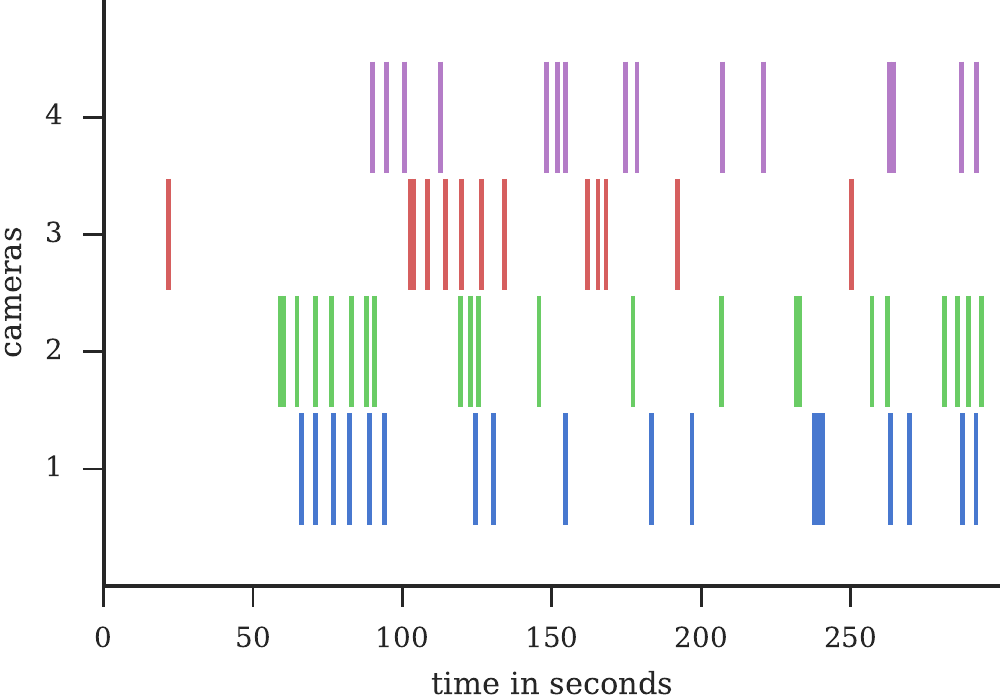}\caption{\label{fig:events_not_synchronized}Flashes detected
in cameras 1-4. The temporal position of the \textit{events} is always in the
camera specific time. The inter camera shift is clearly visible unlike
the clock drift. The drift error accumulates slowly and is not noticeable in this
visualization.}
\end{figure}

\noindent The subsection~\ref{subsec:Abrupt-Light-Changes} and Algorithm~\ref{alg:event_detection} describe the method to detect \textit{synchronization events}. We processed the first 5 minutes of the ice hockey match
in the four video streams and detected 18, 22,
13 and 15 flashes in the cameras 1, 2, 3 and 4 respectively. For the sake of simplicity, we omitted the flashes that crossed the frame boundary. The \textit{event}
distribution is depicted in Figure~\ref{fig:events_not_synchronized}.

We performed two experiments, first we synchronized four cameras jointly by solving a single system of equations, secondly we synchronized camera pairs independently. The results are presented in Table \ref{tab:results_joint} and Table \ref{tab:results_pairs}. The deviation of the synchronized time from the reference time for the detected \textit{events}, which in ideal case should be 0, can be interpreted as a measure of method accuracy. The standard deviation of the synchronization errors is \unit[0.5]{ms} for the joint synchronization and in range from \unit[0.3]{ms} to \unit[0.5]{ms} for the camera pairs. We can claim that our method is sub-millisecond precise.

\begin{figure}
\begin{centering}
\includegraphics[width=1\columnwidth]{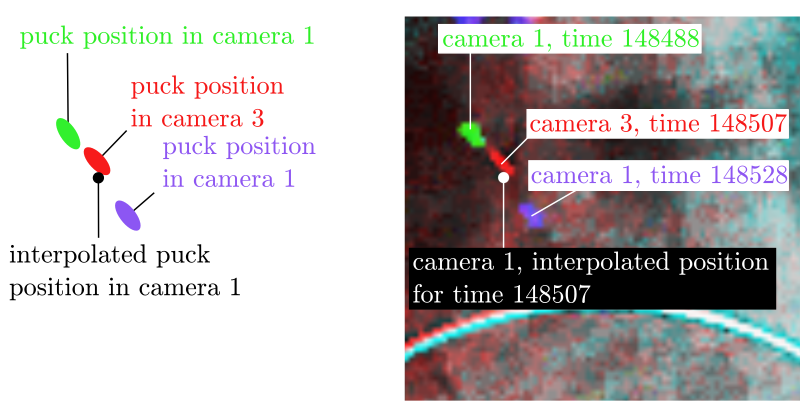}
\par\end{centering}
\caption{\label{fig:puck_interpolation}Synchronization validation. Moving blurred puck is visible in two synchronized cameras. We show three overlaid images of the same puck: two consecutive frames in the camera 1 and single frame in the camera 3. The acquisition time of the puck for all 3 frames was computed considering frame $f$ and row $r$ of the puck centroid. Knowing the puck acquisition times it possible to interpolate a position of the puck in the camera 1 for the time of acquisition in the camera 3. The interpolated puck position in the camera 1 and the real position in the camera 3 should be equal. The situation is redrawn on the left, on the right are the image data visualized in green and blue channels for the camera 1 and in the red channel for the camera 3. The interpolated position in the camera 1 is depicted as a black circle on the left and a white circle on the right. The interpolated position and the real position in the camera 3 are partially overlapping.}

\end{figure}

We validated the found sub-frame synchronization with an observation
of a high-speed object in overlapping views. A puck is present in
two consecutive frames in the camera 1 and in the time between in the camera
3. We interpolated the puck position in the camera 1 to the time of
the puck in camera 3. The puck position in the camera 3 and the interpolated position should
be the same. Figure \ref{fig:puck_interpolation} shows that the
interpolated puck position is close to the real one from camera 3.

\begin{table*}
\caption{\label{tab:results_joint}Synchronization parameters and errors for a system of four cameras. The camera 1 was selected as the reference camera $c_{\mathrm{ref}}$ and the cameras 2, 3 and 4 were synchronized to the reference camera time. The found parameters of the synchronization transformations (Eq. \ref{eq:transformation}) are presented in the table below. The \textit{time per image row} for the reference camera is \unit[0.0154]{ms}. The clock drift is in column three presented as a number of rows per second that need to be corrected to maintain synchronization. The standard deviation of the synchronized time from the reference time for the \textit{synchronization events} is presented in the last column.  }
\centering{}%
\newcolumntype{d}[1]{D{.}{.}{#1} }
\begin{tabular}{r r d{2} d{2} d{4} d{2}}
\toprule 
camera $c_{\mathrm{ref}}$, $c$  & 
1 - clock drift & 
\multicolumn{1}{c}{drift (in $\nicefrac{\mathrm{lines}}{\mathrm{s}}$)} &
\multicolumn{1}{c}{shift (in ms)} &
\multicolumn{1}{c}{$T^c_{\mathrm{row}}$ (in ms)} & 
\multicolumn{1}{c}{std error (in ms)}\tabularnewline
\midrule

 1 2 & 8.39 $\times10^{-6}$  & -0.56 &   6066.7 & 0.015  & 0.49 \\
 1 3 & -3.12 $\times10^{-6}$ &  0.08 & -37500.2 & 0.0394 & 0.44 \\
 1 4 & -8.35 $\times10^{-6}$ &  0.2  & -23858.7 & 0.0414 & 0.44 \\
 
\bottomrule
\end{tabular}
\end{table*}

\begin{table*}
\caption{\label{tab:results_pairs}Synchronization parameters and errors for independent pairs of cameras. For detailed description see Table \ref{tab:results_joint}.}
\centering{}%
\newcolumntype{d}[1]{D{.}{.}{#1} }
\begin{tabular}{r r d{2} d{2} d{4} d{4} d{2}}
\toprule 
$c_{\mathrm{ref}}$, $c$  & 
1 - clock drift & 
\multicolumn{1}{c}{drift (in $\nicefrac{\mathrm{lines}}{\mathrm{s}}$)} &
\multicolumn{1}{c}{shift (in ms)} &
\multicolumn{1}{c}{$T^{c_{\mathrm{ref}}}_{\mathrm{row}}$ (in ms)} & 
\multicolumn{1}{c}{$T^c_{\mathrm{row}}$ (in ms)} & 
\multicolumn{1}{c}{std error (in ms)}\tabularnewline
\midrule

 1 2 & 8.47 $\times10^{-6}$   & -0.57 &   6067.49 & 0.0159 & 0.0148 & 0.45 \\
 1 3 & -8.55 $\times10^{-6}$  &  0.22 & -37500.7  & 0.0158 & 0.0396 & 0.42 \\
 1 4 & -7.04 $\times10^{-6}$  &  0.17 & -23859    & 0.0151 & 0.0417 & 0.39 \\
 2 3 & -14.52 $\times10^{-6}$ &  0.37 & -43567.9  & 0.0149 & 0.0397 & 0.39 \\
 2 4 & -17.37 $\times10^{-6}$ &  0.42 & -29926    & 0.015  & 0.0416 & 0.33 \\
 3 4 & -10.12 $\times10^{-6}$ &  0.2  &  13642.3  & 0.0477 & 0.05   & 0.27 \\
 
\bottomrule
\end{tabular}
\end{table*}

We implemented the system in Python with help of the NumPy, Matplotlib
and Jupyter packages \cite{Hunter2007,Perez2007,VanderWalt2011}.

\section{CONCLUSIONS }

\noindent We have presented and validated a sub-frame time model and a synchronization method for the
rolling shutter sensor. We use photographic flashes as sub-frame synchronization
\textit{events} that enable us to find parameters of an affine synchronization model. The differences of the synchronized time at \textit{events} that should be ideally $0$ are in range from 0.3 to 0.5 milliseconds. We validated the synchronization method by interpolating a puck position between two frames in one camera and checking against the real position in other camera.

We published\footnote{\url{http://cmp.felk.cvut.cz/~smidm/flash_synchronization}} the synchronization code as an easy to use Python module and the paper itself
is available in an executable form that allows anybody to reproduce
the results and figures.

\section*{\uppercase{Acknowledgements}}

\noindent Both authors were supported by SCCH GmbH under Project 830/8301434C000/13162. Jiří Matas has been supported by the Technology Agency of the Czech Republic research program TE01020415 (V3C -- Visual Computing Competence Center). We would like to thank Amden s.r.o. for providing the ice hockey video data.

\bibliographystyle{apalike}
\bibliography{visapp2017}

\end{document}